\documentclass{llncs}

\usepackage[T1,T2A]{fontenc}
\usepackage[utf8]{inputenc}
\usepackage[ukrainian,english]{babel}
\usepackage{graphicx}
\usepackage{booktabs}
\usepackage{amsmath}
\usepackage{amssymb}
\usepackage{cite}
\usepackage{hyperref}
\usepackage{placeins}

\begin{document}
\selectlanguage{english}

\title{Diffusion-Based Ukrainian Handwritten Text Generation with Cross-Domain Style Transfer}

\author{Andrii Ahitoliev\inst{1} \and Pavlo Berezin\inst{2}}
\institute{
Ukrainian Catholic University, Lviv, Ukraine\\
\email{ahitoliev.pn@ucu.edu.ua}
\and
National University of ``Kyiv-Mohyla Academy'', Kyiv, Ukraine\\
\email{pavlo.berezin@gmail.com}
}

\maketitle
\pagestyle{empty}

\begin{abstract}
Handwritten text generation (HTG) conditioned on writer style has been widely studied for Latin scripts, but remains underexplored for low-resource and non-Latin writing systems, leaving open how well existing models generalise beyond the Latin domain. Cyrillic, particularly Ukrainian, lacks both large-scale writer-labeled datasets and empirical evidence of such generalisation. To address this gap, we construct a Ukrainian handwritten word dataset of 126,177 images from 308 writers using connected-component segmentation, quality filtering, and targeted oversampling of underrepresented Ukrainian characters. We retrain DiffusionPen, a MobileNetV2 triplet-loss style encoder with a CANINE-conditioned latent diffusion U-Net, on this dataset without architectural modification, testing direct transfer from Latin to Cyrillic. We evaluate cross-domain style transfer in three settings: cross-lingual transfer from IAM English samples, zero-shot transfer to an early 20th-century Ukrainian manuscript, and few-shot imitation of contemporary writers. The model produces legible, style-consistent word images, indicating that few-shot latent diffusion models generalize beyond the Latin-script domain. We release the dataset, trained models, and evaluation protocol as a reproducible benchmark for writer-aware Cyrillic HTG, providing a foundation for extending stylized HTG to other underrepresented writing systems.

\keywords{Handwritten text generation \and Diffusion models \and Writer-aware diffusion \and Low-resource handwriting}
\end{abstract}

\section{Introduction}

Handwritten text generation (HTG) conditioned on both text content and writer style has advanced substantially in recent years, with diffusion-based methods setting a strong baseline for visually realistic and style-consistent synthesis. However, most of this progress has been demonstrated on Latin-script benchmarks, especially English, leaving open whether the same modeling assumptions transfer to low-resource and non-Latin writing systems. For writer-aware HTG, this is not a minor extension: moving across scripts changes the character inventory, stroke patterns, and available supervision, and current evidence of such transfer remains limited.

Ukrainian Cyrillic provides a meaningful test case for this question. Despite the practical importance of Ukrainian handwriting for digitization and document analysis, the field lacks a large writer-aware word-level dataset suitable for modern HTG and, more importantly, lacks empirical evidence that few-shot style-conditioned generation can generalize beyond the Latin-script domain. As a result, it is still unclear whether recent HTG models capture script-specific regularities only, or whether they learn a more transferable representation of handwriting style.

This paper studies that transfer question through DiffusionPen, a few-shot latent diffusion model originally developed for Latin-script handwriting generation. Rather than treating Ukrainian as only a new training corpus, we use it to test whether a writer-aware diffusion model with style conditioning from reference samples can retain its core behavior when moved to a different script. The key point is not simply that the model can be retrained on Cyrillic data, but that it can do so without architectural modification while preserving meaningful style transfer behavior.

To make that evaluation possible, we construct a Ukrainian handwritten word-level dataset with 126,177 samples from 308 writers. The dataset is derived from line-level source material using connected-component segmentation with 95.7\% boundary accuracy, followed by quality filtering and balancing procedures designed to support generation of underrepresented Ukrainian characters. We then retrain DiffusionPen on this dataset without changing its architecture, allowing the paper to test direct transfer from the Latin-script setting to Ukrainian Cyrillic under a controlled adaptation scenario.

Our evaluation focuses on whether the learned style representation remains useful across domain shifts rather than only on in-domain generation. We therefore study cross-domain style transfer in three settings: cross-lingual transfer from IAM \cite{iam} English handwriting, zero-shot transfer from an early twentieth-century Ukrainian manuscript, and few-shot imitation of unseen contemporary writers. Taken together, these settings provide evidence that writer-aware few-shot latent diffusion models can generalize beyond the Latin-script domain and offer a viable foundation for HTG in underrepresented writing systems.

\textbf{Contributions:}
\begin{itemize}
\item A Ukrainian handwritten word-level dataset with 126K samples from 308 writers.
\item Adaptation of DiffusionPen to Cyrillic without architectural modification.
\item Empirical evaluation of cross-domain style transfer (cross-lingual, historical, few-shot).
\item Analysis of dataset construction factors affecting generation quality.
\end{itemize}

\section{Related Work}

Early handwritten text generation (HTG) systems were dominated by GAN-based approaches conditioned on text content and writer identity. ScrabbleGAN~\cite{scrabblegan} showed that word-level handwriting synthesis could be made length-flexible and useful for downstream recognition, while GANwriting~\cite{ganwriting} introduced reference-based style conditioning from a small set of target-writer samples. Alonso et al.~\cite{alonso2019adversarial} demonstrated sequence-conditioned adversarial generation as a route to recognition augmentation. Subsequent models such as HiGAN~\cite{higan} and Handwriting Transformers~\cite{vatergan} improved style control or long-range consistency, but the broader GAN-based line of work remained limited by training instability and, in many cases, by closed-set assumptions about writer identity.

Transformer architectures have also been applied to HTG. VATr~\cite{vatr} represents target text as sequences of visual archetypes---binary character images rendered in a Unicode font---enabling the model to exploit character-level visual priors for rare and unseen glyphs. VATr++~\cite{vatrpp} extends this with improved style preparation and rare-character augmentation strategies, achieving strong results on IAM. The visual archetype representation is particularly appealing for scripts with large or underrepresented character inventories, a consideration directly relevant to Ukrainian Cyrillic.

Diffusion-based models shifted the field toward more stable training and stronger generation quality. WordStylist~\cite{wordstylist} was the first latent diffusion HTG model to show competitive styled word generation, but it still represented writer style through a discrete writer index, which limits practical transfer to unseen writers. DiffusionPen~\cite{diffusionpen} addressed this limitation by replacing the discrete style label with a few-shot continuous style embedding computed from reference handwriting samples. Subsequent diffusion-based work has further expanded the design space: One-DM~\cite{onedm} reduces the reference requirement to a single image via one-shot diffusion mimicry; DiffBrush~\cite{diffbrush} scales the approach to text-line generation, addressing word spacing and inter-word coherence; and DOG~\cite{dog} proposes a model-agnostic dual orthogonal guidance strategy applied at test time to improve content clarity without retraining. These advances show that the latent diffusion framework remains a productive foundation for ongoing improvement.

Work on HTG for non-Latin scripts remains comparatively sparse. Research has primarily targeted Arabic~\cite{arabicgan2022} and German~\cite{stylusai}, typically framed as data augmentation for recognition rather than open-set style transfer. Evidence for writer-aware transfer across scripts is limited, and work on Cyrillic or Slavic scripts is absent from the prior literature. For Ukrainian in particular, the main obstacle has not only been model adaptation but also the lack of a suitable writer-labeled word-level dataset, as existing Ukrainian handwriting resources are primarily line-level. This leaves an open question that existing literature does not answer: whether a few-shot latent diffusion handwriting model trained in the Latin-script setting can retain meaningful style transfer behavior when adapted to Ukrainian Cyrillic.

\section{Methodology}
 
\subsection{Dataset Construction}
 
No Ukrainian word-level handwriting dataset with writer labels existed before this work. We construct one from the UkrHandwritten line-level corpus~\cite{ukrhandwritten}, which contains 37,111 line images from 331 writers with sentence-level transcriptions.
 
\paragraph{Pre-segmentation cleanup.}
The source scans contain faint dotted or solid underline artifacts beneath the text. If left uncorrected, these propagate into word crops and are learned by the generator as a style feature, producing synthetic words with spurious horizontal lines. A trained NAFNet \cite{nafnet} restoration network is applied to all 37,111 line images before segmentation, producing a cleaned snapshot that serves as input to all subsequent stages.
 
\paragraph{Segmentation.}
Each line image is binarized with Otsu's\cite{otsu} method and decomposed into connected components using OpenCV\cite{opencv}. Components within 8 pixels horizontally are merged into word groups, bridging broken characters and diacritics. The $N{-}1$ largest remaining inter-group gaps are selected as word boundaries, where $N$ is the ground-truth word count from the transcription. This produces exactly one crop per word. On a 500-line evaluation subset, the method achieves a 95.7\% perfect boundary match rate (both boundaries within 5\, px of ground truth), compared with 71.7\% for vertical-projection segmentation and substantially lower rates for off-the-shelf detectors such as CRAFT~\cite{craft} (2.7 words detected per line vs.\ a ground-truth mean of 4.6).
 
\begin{figure}[t]
    \centering
    \includegraphics[width=\textwidth]{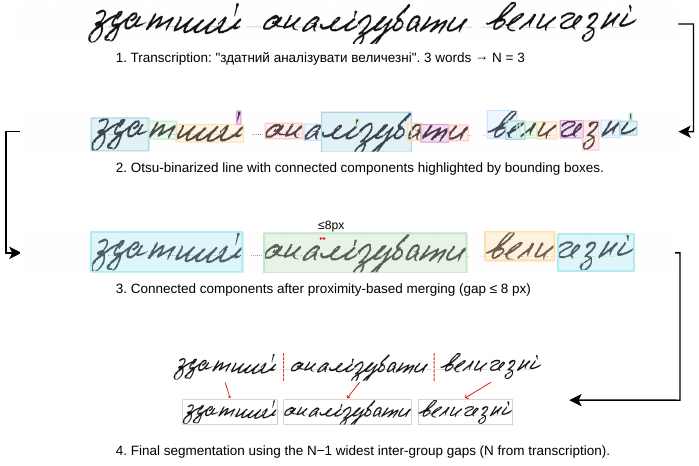}
    \caption{Connected-component word segmentation. Ink blobs within 8\, px are merged into word groups (coloured). The $N{-}1$ widest inter-group gaps (dashed lines) define word boundaries.}
    \label{fig:segmentation}
\end{figure}
 
\paragraph{Filtering.}
Raw segmentation yields ${\sim}$155,000 word crops. A five-stage pipeline removes low-quality samples, with each stage designed to address specific sources of noise or bias: (1)~labels containing Latin characters, pure punctuation, or pure digits are discarded to ensure that the dataset consists solely of valid Ukrainian words, avoiding contamination from irrelevant script and non-lexical items; (2)~crops whose original label ended in a comma are rejected, as these contain a visible trailing comma in the image but not in the normalised label, introducing contradictory supervision that otherwise caused the model to append commas to all generated output; (3)~crops narrower than 20\, px or taller than 100\, px are removed to exclude severely cropped, distorted, or otherwise non-standard samples that could impair the model’s learning of word shapes; (4)~a Cyrillic TrOCR model~\cite{trocr} validates each crop against its label at a similarity threshold of 0.4, with length-stratified relaxation: words of 1--3 characters pass unconditionally, 4--5 characters at threshold 0.2, and 6+ characters at the standard 0.4. This stratification was introduced because the original uniform threshold discarded nearly all short function words, which are frequent in Ukrainian; without these common short words, the model was unable to render words shorter than four characters; (5)~writers with fewer than 50 remaining samples are removed to ensure a minimum level of data per writer, thereby improving the reliability of writer-specific modeling.
 
\paragraph{Balancing.}
Rare Ukrainian letters (\textit{ф}, \textit{ґ}, \textit{Щ}, \textit{Є}, \textit{Ц}, \textit{ї}) are oversampled by duplicating crops containing them 2--5$\times$. The final dataset contains 126,177 word images from 308 writers.
 
\subsection{Model Architecture}
 
\begin{figure}[htbp]
    \centering
    \includegraphics[width=\textwidth]{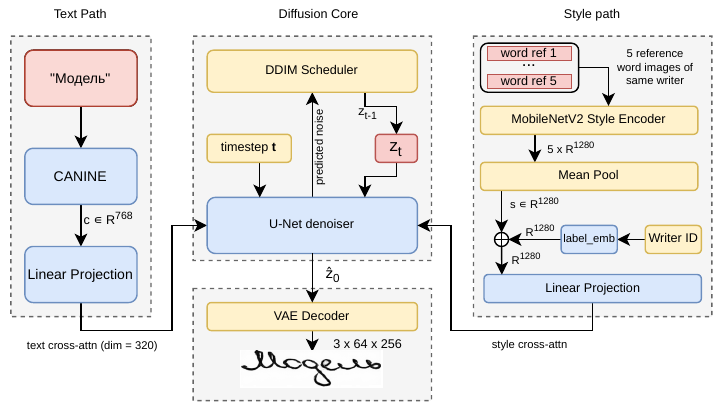}
    \caption{Model architecture at inference.}
    \label{fig:architecture}
\end{figure}

We adopt DiffusionPen~\cite{diffusionpen} without architectural modification. The model is a conditional latent diffusion model operating in the $4{\times}8{\times}32$ latent space of a frozen Stable Diffusion v1.5 VAE~\cite{stablediffusion}. At each denoising step, a U-Net \cite{unet} receives three conditioning signals: (1)~a text embedding $\mathbf{c} \in \mathbb{R}^{768}$ from a CANINE character-level encoder~\cite{canine}, projected to the U-Net dimension of 320; (2)~a style embedding $\mathbf{s} \in \mathbb{R}^{1280}$ from a frozen MobileNetV2~\cite{mobilenetv2} style encoder trained with triplet loss \cite{triplet}, mean-pooled over five reference images; and (3)~a learned writer label embedding summed with $\mathbf{s}$. Both conditioning signals are injected via cross-attention. Figure~\ref{fig:architecture} shows the full architecture.
 
The only configuration change relative to the original DiffusionPen is setting \texttt{num\_res\_blocks\,=\,2} per U-Net level, as specified in Stable Diffusion v1.5. CANINE's Unicode codepoint vocabulary covers Cyrillic natively, requiring no tokenizer modification.
 
\subsection{Training}
 
The model is trained for 200 epochs on the 126K dataset with the standard LDM \cite{ldm} noise-prediction objective. Text conditioning is dropped with probability $p_{\mathrm{drop}}{=}0.2$ for classifier-free guidance (CFG \cite{cfg}); style conditioning is never dropped. In inference, 50 DDIM \cite{ddim} steps with CFG scale $\omega{=}5.0$ are used. Training is performed on a single RTX 4090 GPU with TF32 acceleration and batch size 24.
 
\subsection{Sentence Assembly}
 
Individual word images are assembled into sentence-level strips through three post-processing stages. Baseline alignment shifts each word vertically so that the detected text-body bottom aligns to a common row, using a span-based detector that distinguishes body rows (ink spanning ${\geq}$35\% of image width) from narrow descender strokes. Brightness normalization maps the 5\% of pixels with the lowest brightness in each word to white, ensuring consistent background levels across the strip. Punctuation marks (commas, periods, hyphens) are inserted from a bank of 500 real handwritten marks extracted from the training corpus rather than generated by the model, since the diffusion model does not learn to render isolated punctuation reliably.
 
\begin{figure}[htbp]
    \centering
    \includegraphics[width=\textwidth]{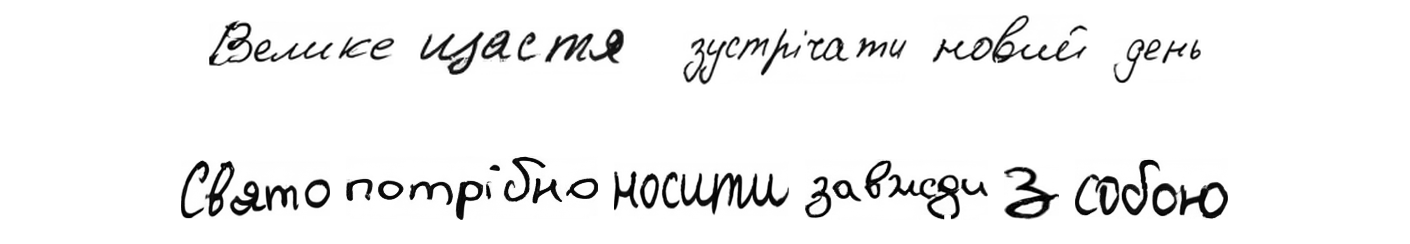}
    \caption{Generated sentences in two writer styles.}
    \label{fig:sentences}
\end{figure}
\section{Experiments}

\subsection{Setup}

All generated images are $64{\times}256$ pixels, matching the training resolution. Quantitative evaluation uses three metrics: Fréchet Inception Distance (FID)~\cite{fid} computed on 5,000 matched writer-word pairs across all 308 writers; Learned Perceptual Image Patch Similarity (LPIPS)~\cite{lpips} computed on the same 5,000 pairs as a secondary pairwise perceptual check; and Character Error Rate (CER) computed by running a pretrained Cyrillic TrOCR model on 4,928 generated words (16 per writer: 8 in-vocabulary, 8 out-of-vocabulary). CER values should be interpreted as an upper bound on generation error, since the Cyrillic TrOCR model itself has non-negligible error rates on real Ukrainian cursive.

\subsection{Quantitative Results}

\begin{table}[htbp]
\centering
\caption{TrOCR Character Error Rate on generated word images.}
\label{tab:cer}
\begin{tabular}{lrr}
\toprule
\textbf{Subset} & \textbf{Samples} & \textbf{CER (\%)} \\
\midrule
Overall & 4,928 & 16.0 \\
\addlinespace
In-vocabulary & 2,464 & 16.5 \\
Out-of-vocabulary & 2,464 & 15.6 \\
\addlinespace
Rare-letter words & 664 & 17.2 \\
Common-letter words & 4,264 & 15.8 \\
\addlinespace
1--3 characters & 1,232 & 42.7 \\
4--6 characters & 1,232 & 10.8 \\
7--9 characters & 1,232 & 12.1 \\
10+ characters & 1,232 & 15.2 \\
\bottomrule
\end{tabular}
\end{table}

\begin{table}[t]
\centering
\caption{FID between real and generated word images.}
\label{tab:fid}
\begin{tabular}{lr}
\toprule
\textbf{Metric} & \textbf{Value} \\
\midrule
FID (5,000 samples, 308 writers) & 23.09 \\
\bottomrule
\end{tabular}
\end{table}

Table~\ref{tab:fid} summarizes visual quality. The FID of 23.09 is comparable to values reported by DiffusionPen on the English IAM dataset (${\sim}$20--25), indicating that the visual quality of generated Ukrainian word images is on par with Latin-script state of the art.

\begin{table}[htbp]
\centering
\caption{LPIPS on the same 5,000 matched real/generated pairs used for FID (AlexNet \cite{alexnet} backbone). Lower is better; handwriting is multimodal, so LPIPS is a secondary indicator.}
\label{tab:lpips}
\begin{tabular}{lrr}
\toprule
\textbf{Subset} & \textbf{Samples} & \textbf{LPIPS} \\
\midrule
Overall mean & 5,000 & 0.367 \\
\addlinespace
1--3 characters & 1,424 & 0.495 \\
4--6 characters & 1,552 & 0.383 \\
7--9 characters & 1,346 & 0.284 \\
10+ characters & 678 & 0.225 \\
\bottomrule
\end{tabular}
\end{table}

Table~\ref{tab:lpips} reports pairwise LPIPS on the same matched pairs. The overall mean of 0.367 is a secondary perceptual indicator: because the benchmark pairs same writer and same transcription by construction, it measures how closely a single generated instance matches one specific real sample, rather than distributional similarity. Handwriting is inherently multimodal---multiple valid renderings exist for the same word and writer---so LPIPS penalizes legitimate stylistic variation and should be read alongside FID rather than in place of it. The length-bucket trend mirrors the CER pattern: longer words yield lower LPIPS (0.225 for 10+ characters) while short words reach 0.495 for 1--3 characters, consistent with the observation that short-word generation is the hardest regime for both legibility and perceptual faithfulness.

Table~\ref{tab:cer} summarizes text legibility. The overall CER of 16.0\% is an upper bound inflated by recognizer limitations, and the writer-macro CER of 16.0\% shows that this performance is not driven only by the most represented writers. The best performance is in the 4--6 character range (CER 10.8\%), where TrOCR is most reliable, while the elevated CER for 1--3 character words (42.7\%) is largely attributable to the recognizer: visual inspection confirms that single-character words are often generated correctly but misread by TrOCR (e.g.,\ ``з'' predicted as ``33'').

Two findings are particularly important. First, OOV words (15.6\%) achieve lower CER than IV words (16.5\%), confirming that the model generalizes to unseen character sequences rather than memorizing training vocabulary. Second, the rare-letter CER gap is only 1.4 percentage points (17.2\% vs.\ 15.8\%), indicating that targeted oversampling substantially narrows the quality gap for underrepresented Ukrainian characters.

\subsection{Contextual Comparison with Prior HTG Systems}

\begin{table}[htbp]
\centering
\caption{Contextual comparison with prior HTG systems ($\downarrow$ lower is better; ``---'' = not reported). Cross-paper values are not directly comparable: datasets, scripts, and evaluation protocols differ. $^{*}$CER from HTR imitation on IAM, different protocol from ours. $^{\dagger}$FID as reported in DiffusionPen (29.94 in WordStylist's own evaluation).}
\label{tab:prior_context}
\begin{tabular}{p{2.3cm}p{1.6cm}rr}
\toprule
\textbf{Model} & \textbf{Dataset} & \textbf{FID}\,$\downarrow$ & \textbf{CER}\,$\downarrow$ \\
\midrule
This paper      & Ukrainian & 23.09 & 16.0\% \\
DiffusionPen~\cite{diffusionpen} & IAM & 22.54 & 6.94\%$^{*}$ \\
WordStylist~\cite{wordstylist}   & IAM & 22.74 & --- \\
GANwriting~\cite{ganwriting}     & IAM & 43.97$^{\dagger}$ & --- \\
\bottomrule
\end{tabular}
\end{table}

Table~\ref{tab:prior_context} places the Ukrainian results in the context of prior HTG systems. The most stable direct comparison available today is FID: our value of 23.09 lies in the same numerical range as IAM-based diffusion HTG reports such as DiffusionPen (22.54). At the same time, these comparisons are contextual rather than strictly like-for-like, because the script, dataset, recognizer, and evaluation pipeline differ, and published FID values vary noticeably even across IAM papers.

The CER values require even more caution. Our CER is an OCR-on-generated-Ukrainian-words measure based on Cyrillic TrOCR, whereas prior English HTG papers often report HTR imitation or augmentation performance on IAM. For that reason, we use the external numbers mainly to position the Ukrainian system within the broader HTG landscape, while treating FID and CER as primary internal baselines for future Cyrillic-to-Cyrillic comparisons.

\subsection{Ablation Study}

Three ablations were conducted on earlier dataset and architecture variants to identify which factors most affect generation quality. All comparisons use validation MSE on a fixed 512-sample held-out set.

\begin{table}[t]
\centering
\caption{Summary of ablations used to select the final training setup.}
\label{tab:ablations}
\begin{tabular}{lrllr}
\toprule
\textbf{Variant} & \textbf{Words} & \textbf{Segmentation} & \textbf{Blocks} & \textbf{Val.\ MSE} \\
\midrule
UltraClean & 12.8K & Projection & 1 & 0.0923 \\
HighConf   & 22.7K & Projection & 1 & 0.0900 \\
Clean      &   67K & Projection & 1 & 0.0787 \\
CC-Clean   &   99K & CC         & 1 & 0.0658 \\
CC-Clean   &   99K & CC         & 2 & ${\sim}$0.063 \\
\bottomrule
\end{tabular}
\end{table}

\paragraph{Dataset size vs.\ purity.}
Three dataset variants were compared: UltraClean (12.8K words, TrOCR@0.75 + edge-ink filter), HighConf (22.7K, TrOCR@0.75), and Clean (67K, TrOCR@0.4). The largest dataset achieved the lowest MSE (0.0787) despite containing substantially noisier samples, outperforming UltraClean (0.0923) and HighConf (0.0900). This confirms that diversity dominates purity for diffusion-based HTG.

\paragraph{Segmentation quality.}
Switching from vertical-projection segmentation (71.7\% boundary accuracy) to connected-component segmentation (95.7\%) on an otherwise identical training setup dropped MSE from 0.0787 to 0.0658, the single largest improvement observed. Misaligned boundaries in the projection-segmented data created systematic label-image inconsistency that CC segmentation largely eliminates.

\paragraph{Residual block count.}
Increasing \texttt{num\_res\_blocks} from 1 to 2 per U-Net level on the same CC-segmented dataset further reduced MSE from 0.0658 to ${\sim}$0.063 and produced visibly sharper glyphs and more reliable diacritic rendering.

\subsection{Generation Quality}

\begin{figure}[htbp]
\centering
\includegraphics[width=\textwidth]{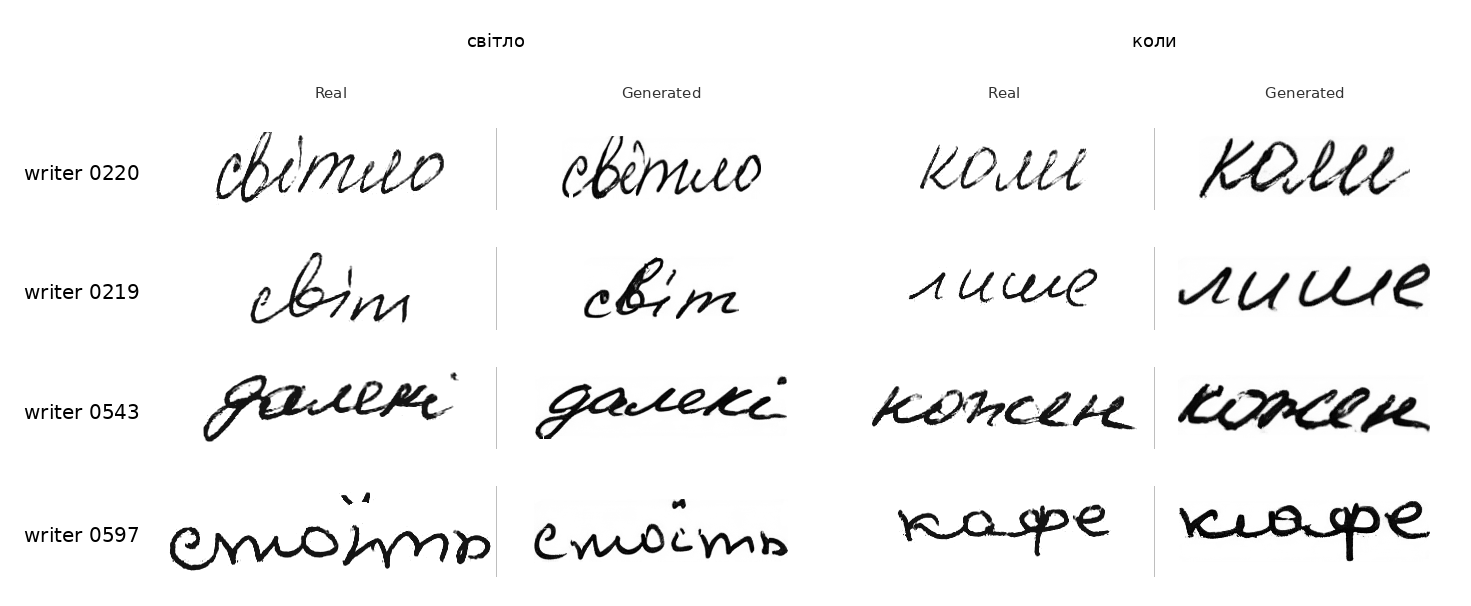}
\caption{Word-level style reproduction on seen writers.}
\label{fig:style_pairs}
\end{figure}

Beyond aggregate metrics, the model must preserve writer-specific style under controlled lexical conditions. Figure~\ref{fig:style_pairs} compares real word crops and generated versions of those same words for the same writers. This same-word layout provides a stricter check than a fixed shared-vocabulary grid, since corresponding letters can be compared directly for slant, stroke endings, spacing, and connectivity.

\subsection{Limitations}

The model is strong on most Ukrainian words and writers, but two edge cases remain. First, rare Cyrillic letters such as \textit{ґ} and infrequent uppercase forms are occasionally substituted or rendered less clearly, although the rare-letter CER gap remains small. Second, apostrophe-bearing words such as \textit{м'яч} and \textit{комп'ютер} remain challenging because the apostrophe is visually tiny, merges with neighboring strokes, and is absent from many writers' samples. These cases affect a limited portion of Ukrainian vocabulary but remain the clearest residual failure modes of the current system.

\section{Cross-Domain Style Transfer}

The triplet-loss style encoder learns a metric space based on visual stroke properties rather than writer identity labels. This makes it possible to extract meaningful style embeddings from handwriting samples entirely absent from training, including samples in other scripts. We test this capability in three settings of increasing domain shift.

\subsection{Cross-Lingual Transfer: English to Ukrainian}

Style reference images are drawn from the IAM Handwriting Database~\cite{iam}, which contains English handwriting from writers with no overlap with the Ukrainian training set. Five reference word images from a single IAM writer are passed through the style encoder, and the resulting embedding is used to generate Ukrainian words. The generated output visibly reproduces the source writer's stroke weight, angle, and spacing.

\begin{figure}[htbp]
    \centering
    \includegraphics[width=\textwidth, height=\textheight, keepaspectratio]{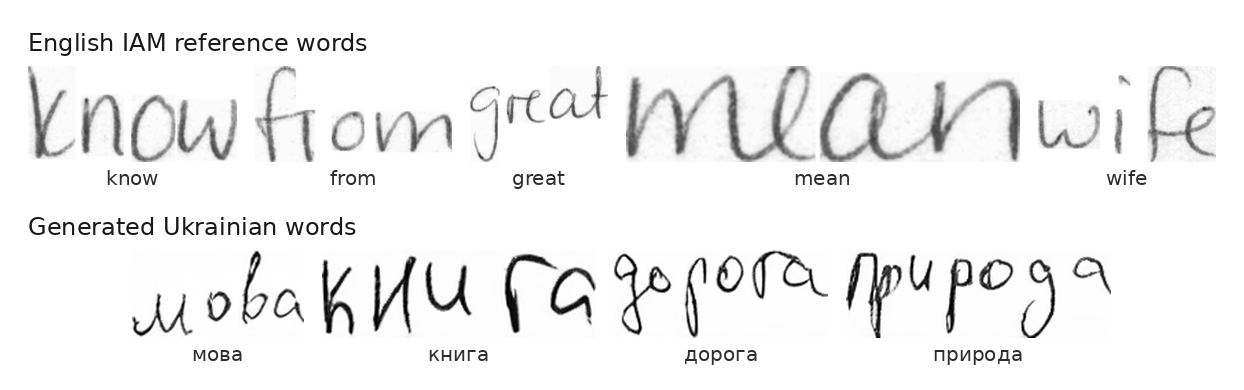}
    \caption{Cross-lingual style transfer.}
    \label{fig:cross_lingual}
\end{figure}

\subsection{Historical Archival Transfer}

Reference images are sourced from a digitized early twentieth-century Ukrainian manuscript archived by the Central State Historical Archives of Ukraine~\cite{archium_manuscript}. These images differ from training data in ink quality, paper texture, and letterform conventions. The generated words adopt the manuscript's calligraphic qualities: wider strokes, more formal letter proportions, and reduced inter-letter connectivity. The model generates modern Ukrainian character forms rather than historical orthography, but the visual style is recognizably derived from the archival source.

\begin{figure}[htbp]
    \centering
    \includegraphics[width=\textwidth]{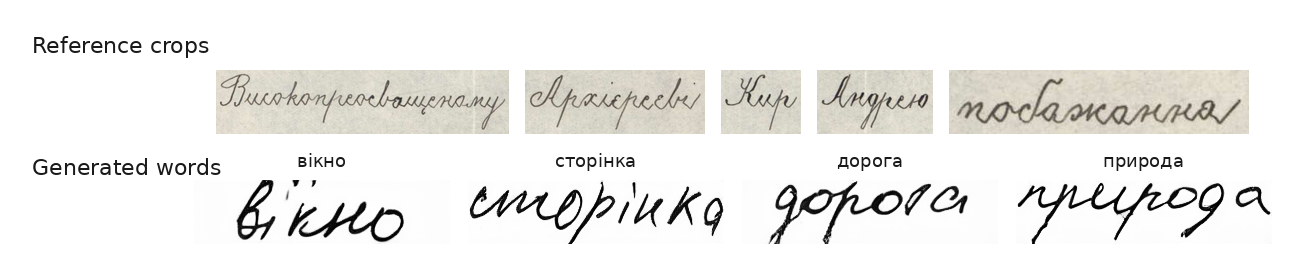}
    \caption{Historical archival style transfer from an early twentieth-century Ukrainian manuscript~\cite{archium_manuscript}.}
    \label{fig:archival}
\end{figure}

\subsection{Unseen Contemporary Writer Transfer}

Reference images are drawn from the RUKOPYS dataset~\cite{rukopys_2026}, a separately collected Ukrainian handwriting corpus whose writers do not appear in the training set. The generated words capture the unseen writer's slant, stroke weight, and letter shapes without any fine-tuning. The quality is comparable to generation for seen writers, confirming that the five-shot style encoding mechanism generalizes to new writers at inference time.

\begin{figure}[htbp]
    \centering
    \includegraphics[width=\textwidth]{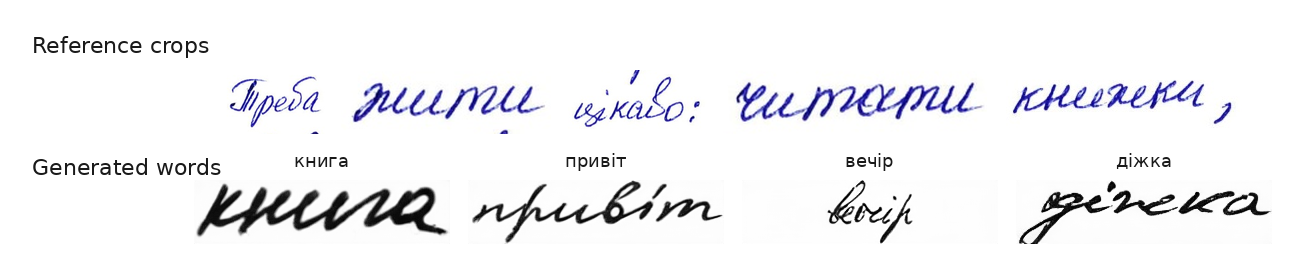}
    \caption{Zero-shot style transfer to an unseen contemporary writer from the RUKOPYS dataset~\cite{rukopys_2026}.}
    \label{fig:modern}
\end{figure}
\FloatBarrier
\section{Conclusion}
 
This paper studied whether few-shot writer-aware latent diffusion models generalize beyond the Latin-script domain. By constructing a 126K-sample Ukrainian word-level dataset from the UkrHandwritten corpus and retraining DiffusionPen without architectural modification, we showed that the model achieves an FID of 23.09 (comparable to its Latin-script performance) and produces legible output across the full Ukrainian character set, with a TrOCR CER of 0.108 on medium-length words.
 
The cross-domain style transfer experiments provide the central evidence: style embeddings extracted from English handwriting, a historical Ukrainian manuscript, and unseen contemporary writers all produce visually style-consistent Ukrainian output. This indicates that the triplet-loss style encoder captures script-independent writing properties and that the CANINE text encoder handles Cyrillic without modification.
 
Training ablations showed that dataset size dominates purity, segmentation quality is the single largest factor in generation quality, and matching the U-Net residual block count to the original Stable Diffusion specification provides a further consistent improvement. 
 
Remaining limitations include reduced fidelity for rare Ukrainian character that occur infrequently even after oversampling, and unreliable rendering of the Ukrainian apostrophe, which affects fewer than 1\% of training samples. Future work includes synthetic augmentation for rare characters, line-level generation to improve sentence coherence, and extension to other Cyrillic-script languages.

\bibliographystyle{splncs04}
\bibliography{references}

\end{document}